\pdfoutput=1
\documentclass[11pt]{article}

\usepackage{coling}

\usepackage{times}
\usepackage{latexsym}

\usepackage[T1]{fontenc}
\usepackage[utf8]{inputenc}

\usepackage{microtype}

\usepackage{inconsolata}

\usepackage{graphicx}

\usepackage[frozencache,cachedir=.]{minted}
\usepackage{multirow}
\usepackage{colortbl}
\usepackage{booktabs}
\usepackage{placeins}
\usepackage{float}
\usepackage[inline]{enumitem}
\usepackage{diagbox}
\usepackage{cleveref}
\usepackage{hyperref}

\usepackage{setspace}
\usepackage{caption}
\title{CarMem: Enhancing Long-Term Memory in LLM Voice Assistants through Category-Bounding}

\author{
 \textbf{Johannes Kirmayr\textsuperscript{1,2}},
 \textbf{Lukas Stappen\textsuperscript{1}},
 \textbf{Phillip Schneider\textsuperscript{3}},
 \textbf{Florian Matthes\textsuperscript{3}},
 \\
 \textbf{Elisabeth André \textsuperscript{2}}
\\
\\
 \textsuperscript{1}BMW Group Research and Technology, Munich, Germany
 \\
 \textsuperscript{2}Chair for Human-Centered Artificial Intelligence, University of Augsburg, Germany
 \\
 \textsuperscript{3}Chair for Software Engineering for Business Information Systems,
 \\
 Technical University of Munich, Germany
\\
 \small{
   \textbf{Correspondence:} \href{mailto:email@domain}{Johannes.Kirmayr@bmwgroup.com}
 }
}

\begin{document}
\maketitle

\newcommand{\lc}[1] {{\color{orange} {{Lukas: #1}}}} % Lukas
\newcommand{\sa}[1] {{\color{NavyBlue} {{\textbf{Philip:} #1}}}} % 
\newcommand{\sae}[1] {{\color{PineGreen} {{#1}}}} % Shahin edit

\newcommand{\cm}{\textsc{CarMem\,}}

\newcommand{\eg}{e.\,g., }
\newcommand{\ie}{i.\,e., }
\newcommand{\wrt}{w.\,r.\,t.\ }
\newcommand{\et}{{et al.\ }}
\newcommand{\cf}{{cf.\ }}

\begin{abstract}
% This document is a supplement to the general instructions for COLING 2025 authors. It contains instructions for using the \LaTeX{} style files for COLING 2025.
% The document itself conforms to its own specifications, and is therefore an example of what your manuscript should look like.
% These instructions should be used both for papers submitted for review and for final versions of accepted papers.

% In this work, a long-term memory system for assistants is proposed that is bound to user pre-defined categories. 
% Idee 1: Personalization in conversational virtual assistants depends on effective long-term memory systems. 
In today's assistant landscape, personalisation enhances interactions, fosters long-term relationships, and deepens engagement. 
However, many systems struggle with retaining user preferences, leading to repetitive user requests and disengagement.
Furthermore, the unregulated and opaque extraction of user preferences in industry applications raises significant concerns about privacy and trust, especially in regions with stringent regulations like Europe.
In response to these challenges, we propose a long-term memory system for voice assistants, structured around predefined categories.
This approach leverages Large Language Models to efficiently extract, store, and retrieve preferences within these categories, ensuring both personalisation and transparency. 
We also introduce a synthetic multi-turn, multi-session conversation dataset (\cm), grounded in real industry data, tailored to an in-car voice assistant setting. 
Benchmarked on the dataset, our system achieves an F1-score of $.78$ to $.95$ in preference extraction, depending on category granularity. Our maintenance strategy reduces redundant preferences by $95\%$ and contradictory ones by $92\%$, while the accuracy of optimal retrieval is at $.87$. Collectively, the results demonstrate the system's suitability for industrial applications.
% for the most detailed category-based preference extraction, with higher performance achieved in broader categories. Additionally, we observe an accuracy of $.80$ in filtering irrelevant preferences outperforming traditional methods in both preference maintenance and retrieval.

% 90% fewer redundant preferences as equal pref- 526
% perences were passed, and 92% less contradictory 527
% preferences as negated preferences are updated

% .86 if k=1

% --> results indicate suitable for industry use
\end{abstract}

\section{Introduction}

\begin{figure*}[t]
  \centering
  \includegraphics[width=.8\textwidth]{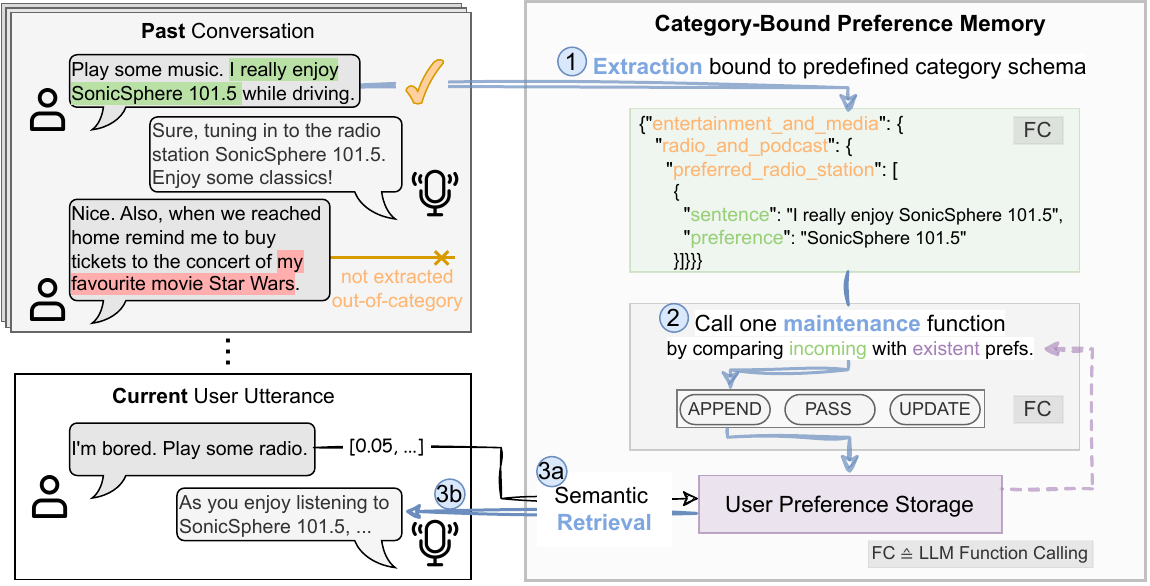}
  \caption{High-level memory flow: 
  After a conversation, preferences are extracted (1) based on the predefined category schema (e.g. preferred radio station). Topics outside the category schema, such as favourite movies, are not extracted. 
  %Preference is extracted from a past completed conversation bounded to the predefined categories: while the preference for Italian food is extracted in a structured JSON, the favourite football club is not extracted as it is out of the categories. 
  (2) Before inserting a new preference, it is compared to existing preferences for consistency, applying the most suitable maintenance operation: append, pass, or update.
  %it gets compared to the existing preferences to keep it up-to-date. One of the three maintenance functions is called based on the situation.
  Within the next conversation (3), the voice assistant retrieves semantically relevant preferences (3a) from the user storage (3b) to provide a personalized response. 
  }
  \label{fig:architecture}
  \vspace{-0.35cm}
\end{figure*}

%The ability to retain and recall memories is fundamental for human interaction as it facilitates the formation of long-term relationships 
Memory retention is essential in human interaction for building long-term relationships
\cite{alea2003whyare, brewer2017howtoremember}. 
Similarly, virtual dialogue systems aim to leverage conversation memories for a more personalised user experience. 
Large Language Models (LLMs) have become a prominent technology in powering such virtual dialogue systems. 
Given that LLMs are inherently stateless, all relevant memories need to be presented during each interaction. 
% For this, it is not practical to present all past interaction messages directly to the LLM with each call, as LLMs degrade in performance \cite{liu-etal-2024-lost} and become more costly with extended input lengths. 
Presenting all past messages to an LLM degrades performance \cite{liu-etal-2024-lost} and increases costs.
Therefore, an external preference memory system is needed that selectively presents a relevant subset of previously extracted memories for the current conversation turn.
However, when engaging with virtual non-human assistants like an in-car personal voice assistant, limitations and concerns arise: 

(1) Privacy Concerns: End-users may have concerns about the extraction and storage of private information from their interactions. 
In Europe, the GDPR \cite{gdpr} enforces data minimization, requiring that data be "adequate, relevant, and limited to what is necessary" for the purposes it is processed under Article 5(1)(c).
Additionally, the EU AI Act \cite{euAIact} mandates a high degree of transparency, reinforcing the need for clear communication about how user data is handled.
(2) Technological Constraints: In-car voice assistants are limited in the information they can actually use due to the restricted action space of the vehicle’s systems.
% in the information they can actually use by their subsequent functional capabilities within the vehicle. 
For example, the preferred radio station can be set as a parameter in the entertainment system, while the favourite movie genre is not applicable.
% While the need for handicapped parking can be set as a paramter in the navigation search, the user's favourite animal is not applicable.
% For example, the user favourite cuisine can be used w
Unbounded information extraction would lead to irrelevant and resource-inefficient storage of memories. 

Our work addresses these industry-relevant challenges by proposing a category-bound preference memory system. This system restricts information extraction, with a focus on user preferences, to hierarchically predefined categories. 
%It thereby avoids the capture of unusable information and allows end-users to control the category topics to which they consent for preference extraction.
Thereby, companies pre-define categories to prevent capturing non-actionable information, and users have the control to further refine this by opting out of specific categories.
An overview of the category-bound preference memory flow is shown in Figure \ref{fig:architecture}. 
The memory system consists of three main components. 
(1) Extraction, which captures in-category preferences after conversations while ignoring out-of-category ones.
%(1) The preference extraction after a conversation, which extracts preferences contained in the categories while ignoring out-of-category preferences. 
(2) Maintenance based on \citet{bae-etal-2022-keep}, which keeps the preference storage up-to-date by calling a maintenance function before storing a preference. (3) Retrieval, which semantically retrieves relevant preferences for the current user utterance to provide personalized responses.

Furthermore, we introduce a carefully constructed synthetic dataset. This dataset focuses on an in-car voice assistant context with multi-turn interactions. The dataset is designed to evaluate the main components of the external memory system. 
%The dataset is focuses on an in-car voice assistant context and includes realistic multi-turn user-assistant interactions including user preferences from predefined categories as well as new session user utterances for retrieval and maintenance of the extracted preferences. 
%Furthermore, we introduce a carefully constructed synthetic dataset, grounded on data of currently deployed intent-based assistants and real in-car conversations, which targets the evaluation of the external memory system.
%The dataset focuses on an in-car voice assistant context with multi-turn interactions.
We benchmark our system on the dataset.
In summary, the main contributions of this work are:
\begin{enumerate}[noitemsep,topsep=3pt]
    \item Category-bound preference memory system based on user-assistant conversations.
    \item Closed-world in-car conversational dataset \cm with benchmark values for main components of our long-term memory system.
\end{enumerate}
Our dataset and code are publicly available.
%\footnote{See attachment; released on github upon acceptance.}
\footnote{Dataset and Code is available on \href{https://github.com/johanneskirmayr/CarMem}{https://github.com/johanneskirmayr/CarMem}.}
\section{Related Work}\label{sec:related_work}

\begin{figure*}[t]
  \centering
  \includegraphics[width=.92\textwidth]{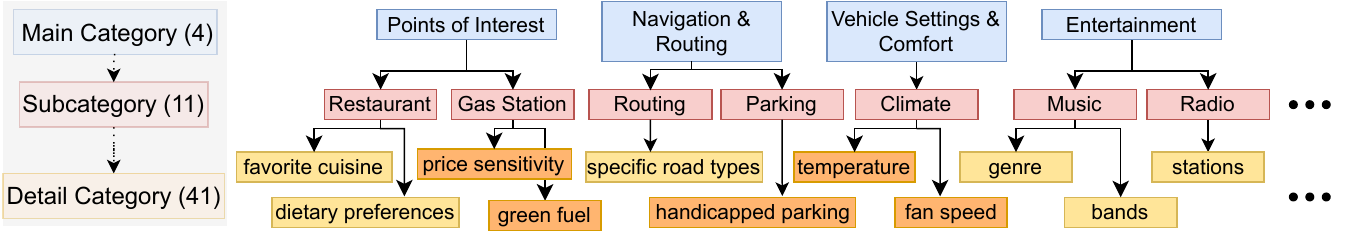}
  \caption[Subset of Manually Defined Hierarchical Preference Categories]{Representative subset of the hierarchically predefined preference categories. There are two types of detail categories: MP (yellow): Multiple preferences within the category are possible, and SP (orange): Single preference within the category is allowed. A full list of categories with attributes is provided in Appendix \ref{subsec:full_list_categories}.}\label{fig:subset_categories}
  \vspace{-0.35cm}
\end{figure*}

Cognitive neuroscience distinguishes between semantic memory (general knowledge) and episodic memory (personal events) \cite{tulving:episem}. While LLMs effectively cover semantic memory, episodic memory must be handled manually. Personalized dialogue systems aim to leverage episodic memory to enhance user experience by tailoring interactions based on individual preferences.
Early approaches used static user profiles \cite{zhang-etal-2018-personalizing}, while more dynamic methods include memory-augmented networks \cite{meng2018}, memory-augmented LLMs \cite{wang2024augmenting}, and external memories that continuously update user memories \cite{xu-etal-2022-beyond, xu-etal-2022-long, xu2023large}. Due to scalability issues with memory-augmented LLMs, we focus on external memory systems that retrieve relevant information as needed.
Several works have explored external memories. \citet{Park2023GenerativeAgents} used an event-based memory in LLM-powered characters for personalized interaction with other characters. 
MemGPT \cite{packer2023memgpt} introduces an operating-system-inspired dual-memory structure. Meanwhile, \citet{Zhong_Guo_Gao_Ye_Wang_2024} enhances its memory mechanism by introducing a human-like forgetting curve.

These advancements, however, have brought new challenges: they deploy unstructured extraction methods, which result in unordered memory pieces in text format, making structured and transparent information an underexplored area. Additionally, with the growing focus on transparency in AI \cite{explainableAIsurvey}, regulations like GDPR \cite{gdpr}, and the EU AI Act \cite{euAIact}, there is increasing demand for systems that offer users more control. 
OpenAI introduced a memory feature in their ChatGPT interface \cite{openAImemory}, where user control is limited to deleting memories after extraction. 
Our approach differs by allowing users to control what gets extracted initially through the ability to opt-out from specific category topics. 
%This approach keeps the intelligence while acting on user privacy wishes.

For maintaining relevant memory, \citet{xu-etal-2022-long} use cosine similarity to remove duplicates, and \citet{bae-etal-2022-keep} introduced LLM-driven memory maintenance. 
We extend this with LLM function calling and structured information representation.
Retrieval-augmented generation based on embeddings \cite{NEURIPS2020_6b493230} has been adapted for preference storage and retrieval \cite{Zhong_Guo_Gao_Ye_Wang_2024, wang2024enhancing}. 
In addition, our system leverages category-based storage to enrich embeddings, improving retrieval accuracy.

These advancements are often limited by the datasets available for evaluation. 
Existing datasets, either focus on user-user conversations \cite{xu-etal-2022-beyond}, are open-domain \cite{xu-etal-2022-long}, or consist of only a single conversational session \cite{zhang-etal-2018-personalizing}. 
Additionally, datasets such as \cite{convAI2} emphasise the assistant's persona rather than user-specific preferences, making them unfit for evaluating long-term, personalised systems, particularly in the context of in-car voice assistants. 
To address these gaps, we introduce a synthetically generated dataset. 
Synthetic datasets have been proven effective in simulating complex, controlled scenarios, especially when real-world data is difficult to obtain \cite{synthetic_computervision, gonzales2023synthetic, wang-etal-2023-target}.

\section{Structured and Category-Bound User-Preference-Memory}

Our system manages user preferences through three stages: hierarchical preference extraction, ongoing maintenance, and retrieval for future interactions.

\subsection{Preference Extraction}\label{subsec:pref_extraction}

% We want to extract user preferences from conversations between the user and the voice assistant. Thereby, the extraction is constrained to pre-defined, hierarchically structured, categories. An illustration of relevant categories for the in-car voice assistant is shown in Fig. \ref{fig:subset_categories}.
Preferences are extracted from conversations and constrained to predefined hierarchical categories. Relevant categories aligned to the in-car assistant are shown in Figure \ref{fig:subset_categories}.
With this, a user could have a preference for Italian food within the category Points of Interest (Main), Restaurant (Sub), Favourite Cuisine (Detail). 
Category-bound extraction (1) increases the transparency by showing which preferences are stored and where; (2) allows users to opt out of categories, for instance, due to privacy concerns; and (3) aligns with the limited action space of downstream car functions, avoiding irrelevant preferences. 
%To achieve the extraction within the categories and its hierarchy we leverage the technique of function calling by an LLM. 
Hierarchical, category-based extraction is achieved via LLM function calling.

\textbf{LLM Function Calling:} %Unlike simple prompt-based extraction methods, function calling allows for greater control and reliability in extracting structured information. 
%There, the LLM is constrained by a predefined parameter schema, which ensures the output adheres to specific format (JSON), minimizing errors. This method reduces ambiguity by ensuring that only relevant information within the designated categories is captured, improving both accuracy and scalability, particularly for industry applications.
Function calling enhances control and reliability in extracting structured information compared to simple prompt-based methods. The LLM is trained to match a predefined parameter schema, ensuring a specific output format (JSON) and extracting only relevant information from the input text for the designated function parameters. 
%As the LLMs are trained to adhere to the schema, this approach minimizes errors and improves both accuracy and scalability, particularly for industry applications.

A function definition consists of the name of the function, a description of the purpose, and a parameter schema. We define a function to extract preferences and use the function parameter schema to represent our categories and their hierarchy as parameters. 
The parameter schema is defined with pydantic \cite{Pydantic} and presented in Appendix \ref{sec:app_preference_extraction}. In the schema, we define every parameter, representing one category (favourite cuisine, preferred radio station, etc.), as \texttt{Optional} so that the LLM is not forced to extract a preference within that category. By using the extraction function on a conversation, the LLM fills in the values of the nested schema, effectively extracting preferences according to the predefined categories and their hierarchy. Out-of-category preferences are either ignored by the LLM as there is no fitting function parameter or extracted in our designated \verb|no_or_other_preference| parameter within the sub- and detail categories which are later discarded. %We intentionally place this parameter before the actual categories to enhance extraction accuracy, given the LLM's token-by-token output processing.

\subsection{Preference Maintenance}\label{subsec:pref_maintenance}
%Once preferences are extracted and categorized, it is essential to maintain the accuracy and relevance of these preferences over time. 
%This can lead to irrelevant or contradictory preferences, negatively impacting user experience and increasing storage costs in large-scale applications. 
%As user preferences change, previously extracted preferences may become outdated or contradictory.
%To ensure they remain up-to-date, we check if the newly extracted preference (incoming preference) is redundant or contradictory to existing preferences before adding it to the database. 
Once extracted, it is essential to maintain the preferences by checking for redundancy or contradictions before storage.
Following \citet{bae-etal-2022-keep}, we have implemented three maintenance functions to account for this:
\textbf{\texttt{Pass}}: The incoming preference already exists in the storage and is not inserted again;
\textbf{\texttt{Update}}: The incoming preference updates an existing preference. The new preference is inserted, and the corresponding existing preference is deleted; 
\textbf{\texttt{Append}}: The incoming preference is new and not present in the storage.
These functions are again used with LLM function calling, defined with a name, description, and parameter schema. The \texttt{append} function requires no parameters, while the \texttt{pass} and \texttt{update} functions need specification of the existing preference causing the call. 
%The LLM is forced to choose one of these functions. 
To streamline comparison, we use the structured storage and present the LLM only existing preferences in the same detail category. Some detail categories allow only a single preference (\cf Figure \ref{fig:subset_categories}) - in these cases, we disable the \texttt{append} function if a preference already exists.

\subsection{Preference Retrieval}\label{subsec:pref_retrieval}
After maintaining an up-to-date database, the next step is to ensure that relevant preferences are retrieved during future interactions.
%The structured extraction allows us to store the user preferences in a tabular database. In addition to storing the preference and its associated categories, we generate an embedding representation of the concatenated string of the detail category, preference attribute, and the sentence revealing the preference. 
To achieve this, we generate an embedding representation from a concatenated string of the detail category, preference attribute, and the sentence revealing the preference.
%Having stored multiple preferences per user from past conversations, we want to retrieve a contextually relevant subset of user preferences based on the current user utterance. We do this by semantic similarity search between the embedding of the utterance and the created preference embeddings.
% Embeddings capture semantic relationships between the user’s preferences and the current context, enabling a more robust and nuanced retrieval process with low latency. This method is particularly effective in situations where the user's utterances vary, as it allows the system to find preferences that are semantically similar, even if the specific wording differs from the original conversation.
Embeddings capture semantic relationships between preferences and context, enabling robust, low-latency retrieval, even with varied user phrasing. We retrieve the most relevant preferences by embedding similarity with the user utterance.
\section{Data}\label{sec:data}

\begin{figure}[t]
    \centering
    \includegraphics[width=\linewidth]{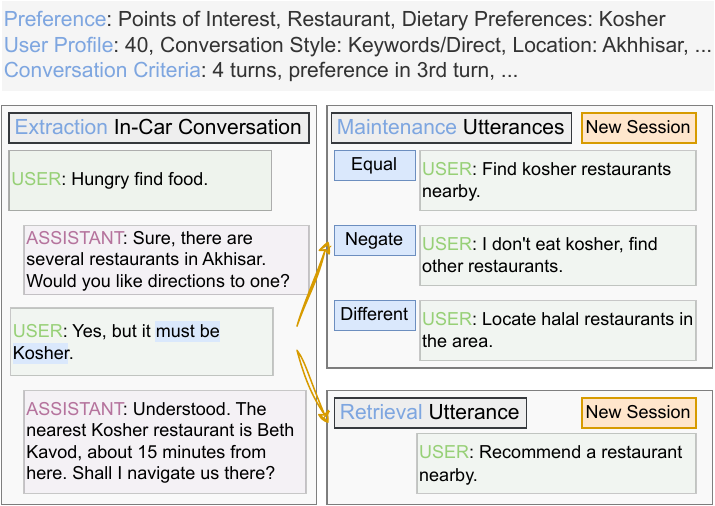}
    \caption{Example data point of the synthetically generated \cm dataset showing the three different parts.}
    \label{fig:example_data_point}
    %\vspace{-0.1cm}
\end{figure}

This section outlines the construction of our synthetically generated dataset \cm. To evaluate the reliability of the category-bound extraction, the dataset features realistic multi-turn in-car \textit{Extraction Conversations} where the user reveals exactly one given preference. Additionally, the dataset includes, in a second session, \textit{Retrieval Utterances} for recalling preferences, and \textit{Maintenance Utterances} for benchmarking maintenance scenarios. Figure \ref{fig:example_data_point} shows an example.

\begin{table}[t]
  \centering
  \begin{tabular}{lr}
    \hline
    \textbf{Statistics} \\
    \hline
    Extraction Conversations & $1,000$ \\
    ~~~~{\small Avg. tokens for generation} & {\small $976$} \\
    ~~~~{\small Avg. turns per conversation} & {\small $5.08$} \\
    ~~~~{\small Avg. words per conversation} & {\small $80.78$} \\
    Retrieval Utterances & $1,000$ \\
    ~~~~{\small Avg. tokens for generation} & {\small $353$} \\
    ~~~~{\small Avg. words per utterance} & {\small $8.34$} \\
    Maintenance Utterances & $3,000$ \\
    ~~~~{\small Avg. tokens for generation} & {\small $357$} \\
    ~~~~{\small Avg. words per utterance} & {\small $12.06$} \\
    \hline
  \end{tabular}
  \caption{Statistics of our \cm dataset.}
  \label{tab:statistics_dataset}
  \vspace{-0.1cm}
\end{table}

To generate the dataset, we use the LLM \verb|GPT-4-1106-preview| \cite{openAImodels} with temperature $0.7$, balancing creativity and coherence (\cf Appendix \ref{sec:app_temperature}). 
To ensure realistic conversations, we prompt the LLM with an elaborate input framework.
For this, we have created 100 user profiles with varying characteristics in age, technological proficiency, user location, and conversation style.
The latter, derived from real-world in-car conversations, ranges from commanding, keyword-only, questioning, to conversational and significantly influences the generated text.
As seen in Figure \ref{fig:example_data_point}, this can result in grammatically incorrect, but realistic interactions.
Each profile is assigned 10 preferences, uniformly sampled across the predefined detail category level (\cf Figure \ref{fig:subset_categories}). 
The categories are based on the most used car functionalities in the currently deployed voice assistant.
For each preference, we create one \textit{Extraction Conversation} where the user reveals the given preference. 
While the user characteristics remain consistent across the 10 generated conversations, the conversation criteria (e.g. conversation length (2-8 turns), position of preference-reveal, preference strength) are randomly sampled for increased diversity (\cf  Appendix \ref{subsec:app_dataset_diversity}). 
Additionally, we provide a real, topic-dependent conversation turn as a few-shot example for each generation.
Strict guidance, random sampling, and the LLM's natural language generation create realistic, controlled, yet diverse dataset entries reflecting preferences relevant to the automotive domain. 
% The resulting dataset comprises 1,000 \textit{In-Car Conversations}, each averaging 5.08 turns and 80.78 words, along with 1,000 \textit{Retrieval Utterances} (8.34 words per utterance) and 3,000 \textit{Maintenance Utterances} (12.06 words per utterance) as detailed in Table \ref{tab}.
The resulting dataset contains 1,000 \textit{Extraction Conversations}, 1,000 \textit{Retrieval Utterances}, and 3,000 \textit{Maintenance Utterances}, detailed statistics are shown in Table \ref{tab:statistics_dataset}. Human evaluation results, showing the dataset's high quality and realism, are in Appendix \ref{subsec:app_dataset_human_evaluation}.
\section{Experiments}

The results are benchmarked on our dataset \cm. We applied a 50-50 split on validation and testing, resulting in 500 test entries. The experiments including an LLM, i.e. extraction and maintenance, were performed using function-calling with the LLM \verb|GPT-4o| (2024-08-06) \cite{openAImodels} at a temperature of 0 to maximize deterministic output (\cf Appendix \ref{sec:app_temperature}).

\subsection{Preference Extraction}\label{subsec:exp_pref_extraction}
We conducted two experiments to evaluate preference extraction from the \textit{Extraction Conversations}: 

\begin{enumerate}[noitemsep,topsep=2pt]
    \item \textbf{In-Schema}: Evaluates if the ground-truth preference can be extracted within the correct categories in the schema. An extraction is considered correct if the main-, sub-, and detail categories match those of the ground-truth preference.
    \item \textbf{Out-of-Schema}: Evaluates if the ground-truth preference is not extracted when the corresponding subcategory is excluded from the schema, simulating a user opt-out. For the example "I want kosher food" the sub-category 'Restaurant' and corresponding detail categories would be excluded from the schema. A data point is considered correct if the ground-truth preference is not extracted.
\end{enumerate}
%he first experiment shows the general extraction capability, while the second highlights the boundness to the categories.

\vspace{-.2cm}
\paragraph{Experiment Setting} Both experiments were conducted on $500$ \textit{Extraction Conversations}, each containing exactly one ground-truth user preference.

The general extraction statistics in Table \ref{tab:exp_extraction_statistics} show a low risk (6\%) of non-extraction when a preference is present and represented in the schema. 
However, when excluding the subcategory from the schema, the non-extraction is desired and achieved 75\% of the time, demonstrating strong boundness to the predefined categories.
In general, we see an incorrect over-extraction with rates of 12\% and 25\%.
The high number of valid structured outputs indicates the reliable adherence to the complex extraction schema, as misformatted JSON outputs and incorrect parameter ($\hat{=}$category) names and hierarchies are labelled as invalid.

\begin{table}[t]
    \centering
    \resizebox{\linewidth}{!}{
    \begin{tabular}{l l r r}
        \hline
        \multicolumn{2}{l}{\textbf{Extraction}} & \textbf{In-Schema} & \textbf{Out-of-Schema}\\ 
        \hline
        no & extraction & 6\% & \textbf{75\%} \\ 
        $1$ & preference & \textbf{82\%} & 23\%\\ 
        $2+$ & preferences & 12\% & 2\%\\ 
        \hline
        \multicolumn{2}{c}{valid struct. output} & 99\% & 99\%\\  
        \hline
    \end{tabular}
    }
    \caption{Statistics for the two \textit{Extraction Conversation} experiments (1) In-Schema and (2) Out-of-Schema, with the ground-truth subcategory included (expects extraction of $1$ preference, highlighted in bold) or excluded in the category schema (expects no extraction, highlighted in bold). The structured output is valid if the output JSON is parseable and matches the schema.}
    \label{tab:exp_extraction_statistics}
\end{table}

\begin{table}[t]
    \centering
    \resizebox{.84\linewidth}{!}{
    \begin{tabular}{r l r r r r}
    \hline
    \textbf{Level} & \textbf{\#cat.} &{\textbf{Prec.}} $\uparrow$ & {\textbf{Rec.}} $\uparrow$ &{\textbf{F1}} $\uparrow$\\ 
    \hline
    Main & 4 & .93 &  .94 & .94 \\
    Sub & 11 & .90 &  .91 & .90 \\ 
    Detail & 41 & .75 & .81 & .78 \\
    \hline
    \end{tabular}
    }
    \caption{\textbf{In-Schema}. Performance scores (micro-averaged) for the \textit{Extraction Conversations} and the ground-truth category included in the category schema. (\#cat.) indicates the number of categories per level.}
    \label{tab:exp_res_extraction}
\end{table}

Table \ref{tab:exp_res_extraction} presents detailed extraction results for the In-Schema experiment. 
While recall for extracting the ground-truth preference and classifying it into the correct main category is high at $.94$, the performance declines with a deeper hierarchy level and an increasing number of categories.
At the most detailed level (41 categories) precision is $.75$, which we see as a crucial score in an industry application, as it is better to not extract a preference than to extract an incorrect one. 
Appendix \ref{subsec:app_cm_detail_cat_in} (Figure \ref{fig:cm_detail}) includes the confusion matrix for the detail level of the In-Schema experiment, showing that most incorrect extractions occur in semantically closely related categories. 
This is further supported by the confusion matrix for the subcategory level of the Out-of-Schema experiment (Appendix \ref{subsec:app_cm_detail_cat_in}, Figure \ref{fig:cm_sub_cat_out}), which shows no incorrect extractions for semantically distinct categories like 'Climate Control' but significantly more confusions for closely related categories like 'Music' and 'Radio and Podcast'. 
These results indicate that defining clear and semantically distinct categories is crucial for achieving reliable category-bound extraction.

\subsection{Preference Maintenance}\label{subsec:exp_pref_maintenance}

Table \ref{tab:res_mapping_query_type_to_function} shows that each of the three \textit{Maintenance Utterance} types is assigned a specific function call as its ground truth label. This mapping is based on the incoming preference from the \textit{Maintenance Utterance}, the existing preference from the \textit{Extraction Conversation}, and the detail category type.
\begin{table}[htpb]
\centering
\small
\begin{tabular}{l l}
\hline
\textbf{Type} & \textbf{Label}\\
\hline
equal preference & $\rightarrow$ \verb|pass| (MP, SP)\\
negate preference & $\rightarrow$ \verb|update| (MP, SP)\\
different preference & $\rightarrow$ \verb|append| (MP) \\
 & $\rightarrow$ \verb|update| (SP)\\
\hline
\end{tabular}
\caption[Mapping of Maintenance Utterance Type to Maintenance Function Label]{Mapping of \textit{Maintenance Utterance} type to maintenance function considering the detail category type (MP: multiple preferences allowed, SP: single preference allowed).}\label{tab:res_mapping_query_type_to_function}
\end{table}
A data point is considered correct if the ground truth maintenance function is called.

\vspace{-.2cm}
\paragraph{Experiment Setting} 
To ensure an independent evaluation, we perform the maintenance evaluation only on the dataset entries with perfect extraction accuracy for both the original preference in the \textit{Extraction Conversation} and the modified preferences in the \textit{Maintenance Utterances}. 
The number of data points for each experiment is shown in Table \ref{tab:exp_maintenance_cm}. 
On average, each user has $7.02$ existing preferences from the corresponding \textit{Extraction Conversations}.

\begin{table}[h]
    \centering
    \resizebox{\linewidth}{!}{
    \begin{tabular}{l r r | c c c}
        %&$\#$ & Type & \multicolumn{3}{c}{Confusion Matrix} \\
        %\hline
         &$\textbf{\#}$& \cellcolor{gray!20} \textbf{Type} & \cellcolor{gray!10} \texttt{pass} & \cellcolor{gray!10} \texttt{update} & \cellcolor{gray!10} \texttt{append} \\
         \cline{3-6}
        \multirow{3}{*}{MP}&159 & \cellcolor{gray!20} equal & \textbf{.86} & .03 & .11 \\ 
        &143 & \cellcolor{gray!20} negate & .00 & \textbf{.87} & .13 \\ 
        &159 & \cellcolor{gray!20} different & .03 & .04 & \textbf{.92} \\
        \hline
        \multirow{3}{*}{SP}&192 & \cellcolor{gray!20} equal & \textbf{.68} & .32 & - \\
        &160 & \cellcolor{gray!20} negate & .02 & \textbf{.99} & - \\
        &192 & \cellcolor{gray!20} different & .01 & \textbf{.99} & - \\
    \end{tabular}
    }
    \caption[Quantitative Results of Maintenance Function Calling]{Modified confusion matrix for the maintenance function calling task, segmented by categories that allow multiple preferences (MP) or a single preference (SP). Expected mapping (highlighted in bold) from \textit{Maintenance Utterance} type to function is shown in Table \ref{tab:res_mapping_query_type_to_function}. (\#) indicates the number of data points used per type.}
    \label{tab:exp_maintenance_cm}
\end{table}
From the weighted average (MP \& SP) in Table \ref{tab:exp_maintenance_cm}, 76\% of equal preferences were correctly passed.
Since updating an equal preference yields the same result as passing it, our maintenance method achieves a 95\% reduction in redundant preferences.
Additionally, contradictory preferences are reduced by 93\% as negated preferences are updated.
However, in 2\%, preferences are still lost due to incorrect passes.
In the MP case, 12\% are still wrongly appended, similar to scenarios without maintenance.

%Taking the weighted average (MP \& SP) from the results shown in Table \ref{tab:exp_maintenance_cm}, 76\% of equal preferences were passed. Since updating a equal preference leads to the same result as passing it, our maintenance methods achieves, on average, 95\% fewer redundant preferences. Moreover, it reduces contradictory preferences by $93\%$ as negated preferences are updated. 
%Taking the weighted average from the results shown in Table \ref{tab:exp_maintenance_cm}, our maintenance method leads to 76\% fewer redundant preferences as equal preferences were passed, and 93\% less contradictory preferences as negated preferences are updated. 
%In 2\% of cases, a preference is lost because it is incorrectly passed. 
%In 12\% of cases, we still wrongly append a preference, as it would happen without the maintenance. 
% While these numbers show the necessity of a maintenance methodology, it could be further improved by iteratively performing maintenance on the existing preference storage as well as taking the temporal aspect into account.

\subsection{Preference Retrieval}\label{subsec:exp_pref_retrieval}

In the \cm dataset, each \textit{Retrieval Utterance} is designed to focus on the topic of the ground-truth subcategory, targeting the retrieval of the corresponding ground-truth preference. While the $k$ for semantic retrieval is fixed in practice, we adapt it dynamically to provide more insightful results. Consequently, retrieval is considered optimal if the ground-truth preference is among the top-$n_{i,j}$ retrieved preferences, where $n_{i,j}$ represents the number of preferences stored for user $i$ within subcategory $j$. On average, the parameter $n$ is $1.57$ and each user has $7.02$ preferences stored.

\vspace{-.2cm}
\paragraph{Experiment Setting} We perform the retrieval experiment on the $351$ preferences with optimal extraction accuracy. 
Embeddings are generated using the OpenAI \verb|text-embedding-ada-002| model.

\begin{table}[htpb]
\centering
\small
\resizebox{\linewidth}{!}{
\begin{tabular}{l r r r r}
\toprule 
 %& & \multicolumn{3}{c}{$\overline{n_s} = 1.57$} \\
%\cline{3-5}
\textbf{Embedding} & \cellcolor{gray!20} $k =$ & \cellcolor{gray!20} \textbf{$n$} & \cellcolor{gray!20} \textbf{$n+1$} & \cellcolor{gray!20} \textbf{$n+2$} \\
\hline
\multicolumn{2}{l}{Sentence only} & .75 & .88 & .93 \\
\multicolumn{2}{l}{Detail Cat.+Attr.+Sent.} & \textbf{.87} & \textbf{.94} & \textbf{.97} \\
\hline
\end{tabular}
}
\caption[Quantitative Results of the Embedding-Based Retrieval Methods]{Top-k accuracy for retrieving the ground-truth preference based on the \textit{Retrieval Utterance}. The parameter $n$ is set dynamically to the number of preferences stored for the user $i$ and subcategory $j$. Embeddings are created either (1) from the sentence where the preference was revealed or (2) enriched by the preference detail category and attribute.}

\label{tab:res_retrieval_performance_metrics_embedding_based}
\end{table}

Table \ref{tab:res_retrieval_performance_metrics_embedding_based} shows the results for two embedding approaches: 
(1) embeddings created solely from the sentence where the preference is revealed, and 
(2) embeddings enriched with the structured extraction, including the detail category and the attribute.
Given that, on average, $7.02$ preferences are stored and $\overline{n}=1.57$, we can observe an effective retrieval. % with accuracy of $.87$ with $22\%$ of the preferences retrieved, and $.97$ with $50\%$ of the preferences retrieved.
Furthermore, the enriched embedding outperforms the 'sentence only' embedding by $.12$ in accuracy for optimal retrieval. 
This improvement is evident in the following example:

\begin{itemize}[itemsep=.5pt, topsep=3pt]
    %\small
    \item \textbf{Sentence only}: "I always find NavFlow to be reliable."
    \item \textbf{Detail Cat.+Attr+Sent.}: "traffic information source preferences: NavFlow. I always find NavFlow to be reliable."
\end{itemize}
We observe that categories clarify ambiguous sentences by providing additional context, and fixed category names help cluster preferences more closely in the embedding space.
\section{Conclusion}\label{sec:conclusion}

We presented a structured, category-bound preference memory system capable of extracting, maintaining, and retrieving user preferences, while enhancing transparency and user control in privacy-critical contexts. 
Our approach utilizes a synthetic dataset grounded in real in-car conversations to ensure realism. Benchmarking the core components of the preference memory on this dataset demonstrated both the system's utility and strong performance. 
Future work could build upon the dataset, refine our baseline methods, and explore generalizing to other industry domains such as smart homes, further validating the approach’s adaptability.
\clearpage

% do not count to page limit
\section{Limitations}\label{sec:limitations}
The dataset contains exactly one preference per conversation, which is beneficial for evaluation but does not account for conversations containing no or multiple preferences. 
While we carefully simulated realistic in-car user-assistant interactions, we did not incorporate additional speech recognition errors or repeated user requests, both of which are common in real-world scenarios. Although LLMs often provide automatic corrections for such issues in practice, structural testing could yield further insights into robustness. 
%The nondeterministic nature of LLMs can introduce errors or unrealistic scenarios. 

Moreover, the dataset represents interactions across only two timeframes, limiting our evaluation to the basic functionalities without testing the long-term ability to adapt to changing user preferences. 
Incorporating techniques such as temporal decay of memorized preferences \cite{Zhong_Guo_Gao_Ye_Wang_2024} or assigning importance ratings\cite{Park2023GenerativeAgents} could improve our maintenance methods.

Although the preference extraction experiment adhered well to the category schema, incorrect over-extraction occurred at rates of 12\% to 15\%. To mitigate this, we propose to leverage in-context learning capabilities of the LLM and provide explicit few-shot examples where no preference should be extracted.
Furthermore, we used OpenAI's JSON mode for data extraction. However, the just-released structured output mode by \citet{openAIstructuredoutput} reportedly adheres 100\% to the provided schema, which could further improve our preference extraction results.

\section{Ethical considerations}\label{sec:ethical}

% example text snippets from one of my own papers, which you can adapt for this study
%In our experiments, we used a self-constructed dialogue dataset while ensuring that no personal identifying information was processed or disclosed. 
%Moreover, to ensure optimal computing efficiency, evaluations were conducted on cloud computing platforms via APIs, with each inference run taking less than X minutes/hours

%\textcolor{red}{to-be-written}
%Our dataset was synthetically generated and does not contain any personally identifiable information. The attributes for categories such as 'favorite artist' or 'preferred radio station' were also generated, ensuring no real individuals or brand names were included. 
%For the user profiles used in dataset generation, we only incorporated neutral information such as age or conversation style, avoiding sensitive attributes like gender or ethnic background. However, since large language models (LLMs) are trained on vast amounts of mostly uncurated online data, they may inherit harmful social biases \cite{10.1162/coli_a_00524}, which could be reflected in our dataset.

Our dataset was synthetically generated and does not contain any personally identifiable information. 
The attributes for the categories such as 'favourite artist' or 'preferred radio station' were also generated, ensuring no real persons or brand names were included.
For the user profiles used in dataset generation, we only incorporated neutral information such as age or conversation style, avoiding sensitive attributes like gender or ethnic background.
However, since LLMs are trained on vast amounts of mostly online data, they inherit harmful social biases \cite{10.1162/coli_a_00524}, which could be reflected in our dataset. By prompting the LLM with bias-neutral few-shot examples, we aimed to guide the model toward fairer extractions.

Our proposed preference memory system is designed to be transparent and explainable in its approach for extracting and managing user preferences. 
This aligns with emerging AI regulations such as the EU AI Act \cite{euAIact} which mandates transparency, and the General Data Protection Regulation (GDPR) \cite{gdpr}, which emphasizes data protection and user consent.
A key aspect of our system is category-bound extraction, which follows the principles of data minimization and user control. By aiming to extract and store only actionable information and allowing users to opt out of specific categories, we preserve user privacy while maintaining system intelligence.

However, despite our system's safeguards, it does not achieve perfect accuracy, and LLMs may hallucinate. This introduces potential risks, such as the extraction of false or irrelevant preferences. To mitigate this, integrating extracted data in the UX flow and transparently displaying them on the user interface, provides users with the ability to manually delete memories.
Additionally, offering an interaction tool via voice  allows users to review, edit, or delete preferences, maintaining system accuracy and trust.
Future work may explore confidence thresholds that trigger user confirmation for uncertain extractions.

\newpage
\bibliography{coling_latex}

\appendix
\clearpage
\section{Prompts}\label{sec:app_prompt_list}
%\textcolor{red}{Prompt List}
The prompts for dataset generation, preference extraction, and maintenance function calling are available in our released code on \href{https://github.com/johanneskirmayr/CarMem}{https://github.com/johanneskirmayr/CarMem}.

\section{LLM Temperature Settings}\label{sec:app_temperature}

The temperature parameter controls the randomness and creativity of the generated text. We used different settings of temperature depending on the task:

\begin{itemize}[itemsep=1pt, topsep=3pt]
    \item \textbf{Dataset Generation:} According to GPT-4 technical report, a temperature of $0.6$ is recommended for free-text generation \cite{openai2024gpt4}. Considering the need for creativity and diversity in dataset generation task, and referencing related work by \citet{wang-etal-2023-target}, which employs a temperature of 0.75, we decided on a temperature setting of $0.7$.
    \item \textbf{Extraction and Maintenance Function Calling:} For the tasks of extraction and maintenance function calling, we set the temperature to $0$. These tasks require precise and consistent outputs without creativity, maximizing deterministic and reproducible results.
\end{itemize}

\section{\textsc{CarMem} Dataset}\label{sec:app_dataset}

\subsection{Human Evaluation}\label{subsec:app_dataset_human_evaluation}

In this section, we present the results of the human evaluation conducted to assess the quality and relevance of the dataset. A subset of 40 data points, systematically selected from 40 users in the \cm dataset, was evaluated by three human judges.
The preferences, which are ordered correspondent to the category list, were chosen in a repeating pattern from the first to the tenth preference. This approach ensured a representative coverage of all preference categories and user profiles. To ensure high intercoder reliability , the judges were provided with detailed instructions. The instructions included the goals for each dataset component, an explanation of the dynamic inputs (user profile, conversation criteria), the evaluation criteria, and guidelines for the different evaluation values. Furthermore, one independent data point was evaluated collaboratively to establish a consistent evaluation standard.

The evaluation criteria for the \textit{Extraction Conversation} part of the \cm dataset are as follows:

\begin{enumerate}[itemsep=1pt, topsep=3pt]
    \item \textbf{Realism of User Behavior}: Does the simulated user behave and communicate in a manner that reflects how real users would act in a similar in-car situation?

    \item \textbf{Realism of Assistant Responses}: Are the assistant's responses contextually appropriate,  relevant, and reflective of a natural understanding of human speech patterns?
    
    \item \textbf{Organicness of User Preference Revelation}: Is the user preference revealed naturally within the flow of the conversation without being forced or out of place?
    
    \item \textbf{Clarity of User Preference}: Is the   user preference communicated clearly, making it distinct from a temporary wish or a one-off statement? 

    \item \textbf{Environment Understanding}: Does the model demonstrate an understanding of the context in which the conversation is taking place?
\end{enumerate}

Each criterion was assessed on a Likert scale from 1 (worst) to 3 (best). 
Additionally, each \textit{Extraction Conversation}, \textit{Retrieval Utterance}, and \textit{Maintenance Utterance} is assessed for appropriateness within the dataset and scored for subjective quality on an overall Likert scale rating (1-3). 
A data point should be scored inappropriate if, for example, the user preference is unclear, the conversation contains multiple preferences, the retrieval utterance already included the ground-truth preference or the maintenance utterances do not fulfil the intended purpose. The majority vote was taken in discordant situations.

\paragraph{Human Evaluation Results} Table \ref{tab:res_human_evaluation_one_pref} details the results of the human evaluation on $40$ datapoints for the \cm dataset.
\begin{table*}[t]
\centering
\footnotesize
%\resizebox{\textwidth}{!}{
    \begin{tabular}{llll}
        \toprule
        \rowcolor{blue!20} \textbf{Criteria} & \textbf{Average Score $[1,3]$$\uparrow$} & \textbf{Ratio 'Appropriate' $[0,1]$$\uparrow$} \\
        \hline
        \rowcolor{gray!20}\multicolumn{3}{l}{Extraction Conversations} \\
        \hline
        Realism of User & $2.73$ & \\
        Realism of Assistant & $2.93$ & \\
        Organicness of User Preference & $2.67$ & \\
        Clarity of User Preference & $2.47$ & \\
        Environment Understanding & $3.0$ & \\
        Overall Subjective Quality & $2.18$ & \\
        Appropriate Conversation for Dataset & & $31/40 = 0.78$ \\
        \midrule
        \rowcolor{gray!20}\multicolumn{3}{l}{Retrieval Utterance} \\
        \hline
        Overall Subjective Quality & $2.75$ & \\
        Appropriate Question for Dataset & & $38/40 = 0.95$ \\
        \midrule
        \rowcolor{gray!20}\multicolumn{3}{l}{Maintenance Utterances} \\
        \hline
        Overall Subjective Quality & $2.71$ & \\
        Appropriate Maintenance Questions for Dataset & & $40/40 = 1.0$ \\
        \bottomrule
    \end{tabular}
%}
\caption{Results of Human Evaluation based on $40$ Data Points of the \cm dataset.}\label{tab:res_human_evaluation_one_pref}
\end{table*}
The results indicate that the \textit{Extraction Conversations} were generally realistic, with high scores in realism and environment understanding. 
The reveal of user preferences was mostly natural and clearly identifiable. 
However, nine conversations were classified as inappropriate: in six cases, the user preferences were not identifiable, and in three cases, multiple preferences, including the ground truth preference, were revealed. 
Both \textit{Retrieval Utterances} and \textit{Maintenance Utterances} showed high overall subjective quality and a high ratio of appropriate utterances. 
For the \textit{Retrieval Utterances}, one instance was classified inappropriate due to the utterance not being related to the user preference, and one because of 'other' reason - here the utterance contradicted the user preference. 
% Since this analysis covers only a subset of the dataset, the scores should be considered approximations. For example, it cannot be definitively concluded that all maintenance utterances in the full dataset are appropriate.

\subsection{Increased Diversity through User Profiles and Conversation Criteria}\label{subsec:app_dataset_diversity}

As detailed in Section \ref{sec:data}, dynamic prompt inputs are sampled for generating each conversation. We hypothesize that this variation in user profiles and conversation criteria will result in increased diversity in the generated text.

\begin{figure*}[t]
  \centering
  \includegraphics[width=0.9\textwidth]{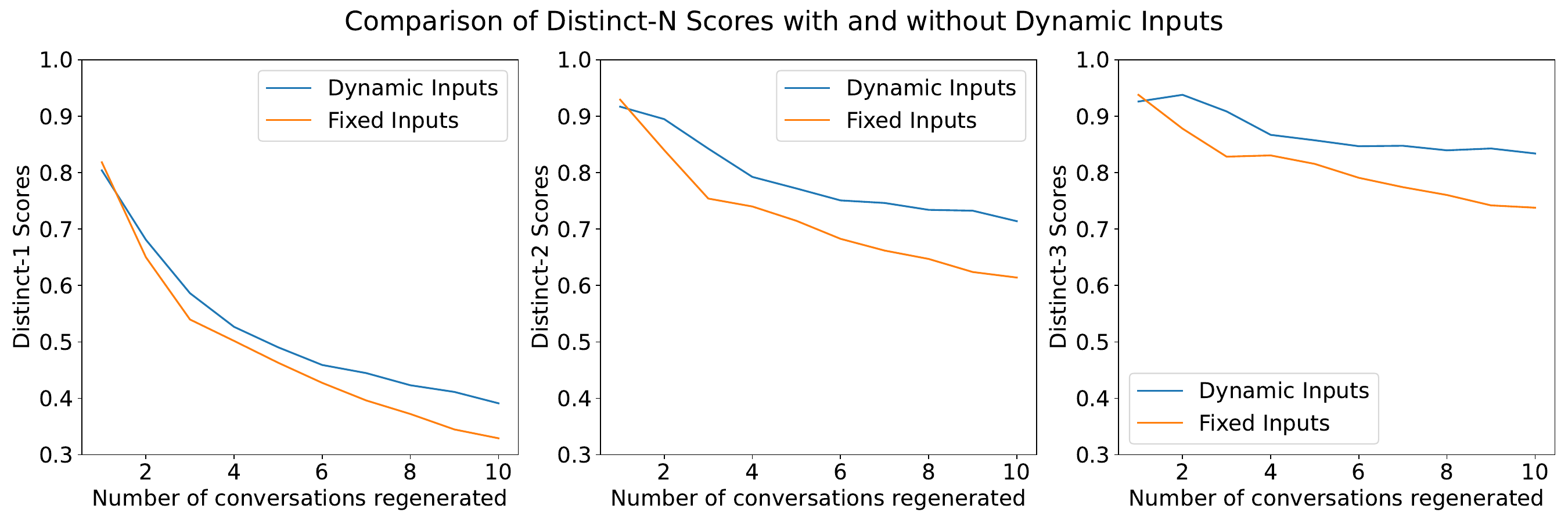}
  \caption[Diversity Evaluation with Dynamic and Fixed Inputs]{The figure shows the diversity evaluation (Distinct-1, Distinct-2, Distinct-3) (y-axis) with dynamic and fixed inputs. The scores were calculated and then averaged for four different user preferences, with each preference's conversations being regenerated 1 to 10 times (x-axis).}\label{fig:res_distinct_n_score}
\end{figure*}

\paragraph{Experiment Setting} To test our hypothesis, we randomly sampled four different user preferences. For each preference, we "regenerate" the conversations 10 times with 2 methods: (1) regenerate with varying dynamic inputs, and (2) regenerate with non-varying fixed inputs. To compare the diversity of the generated conversations, we increasingly concatenate (from 1-10) the regenerated conversations for both methods and calculate the Distinct-1, Distinct-2, and Distinct-3 scores. Calculating the three Distinct-N scores allows for a comprehensive assessment of text diversity across varying levels of lexical and syntactic granularity. The same prompt was used for both methods. The dynamic inputs to the prompt are: User Profile Data (Age, Technological Proficiency, Conversation Style, Location), Conversation Criteria (Position User Preference, Preference Strength Modulation, Level of Proactivity Assistant), and the Few Shot Example. Note: To mitigate the issue of unequal evaluation due to varying text lengths, the conversation length was fixed to six messages for both methods. For the fixed input method, the dynamic inputs were sampled once at the beginning and kept constant across the 10 conversations. For the dynamic input method, inputs were resampled for each conversation. Since the conversation style was found to have a significant influence on the generated text, we exceptionally manually set the conversation style to the four possible values for the four different user preferences in the fixed input method to ensure more representative results. The temperature of each LLM is set to $0.7$.
The averaged Distinct-N scores across the four preference generations can be seen in Figure \ref{fig:res_distinct_n_score}.

As the number of regenerations increases, diversity tends to decrease for both methods. However, the results indicate that conversations with dynamic inputs consistently achieve higher diversity scores across all Distinct-N metrics compared to those without dynamic, but fixed inputs.

\section{Predefined Categories}
\label{sec:pre-defined categories}

\subsection{Full List of Preference Categories with Attributes}\label{subsec:full_list_categories}

In the following, the full list of preference categories with attributes is shown. From this list, every user profile gets sampled 10 preferences.
\begin{enumerate}
\scriptsize
   \item Points of Interest
   \begin{enumerate}
       \item Restaurant
       \begin{enumerate}
           \item MP: Favorite Cuisine
           \begin{itemize}
               \item Attributes: Italian, Chinese, Mexican, Indian, American
           \end{itemize}
           \item MP: Preferred Restaurant Type
           \begin{itemize}
               \item Attributes: Fast food, Casual dining, Fine dining, Buffet
           \end{itemize}
           \item MP: Fast Food Preference
           \begin{itemize}
               \item Attributes: BiteBox Burgers, GrillGusto, SnackSprint, ZippyZest, WrapRapid
           \end{itemize}
           \item SP: Desired Price Range
           \begin{itemize}
               \item Attributes: cheap, normal, expensive
           \end{itemize}
           \item MP: Dietary Preferences
           \begin{itemize}
               \item Attributes: Vegetarian, Vegan, Gluten-Free, Dairy-Free, Halal, Kosher, Nut Allergies, Seafood Allergies
           \end{itemize}
           \item SP: Preferred Payment Method
           \begin{itemize}
               \item Attributes: Cash, Card
           \end{itemize}
       \end{enumerate}
       \item Gas Station
       \begin{enumerate}
           \item MP: Preferred Gas Station
           \begin{itemize}
               \item Attributes: PetroLux, FuelNexa, GasGlo, ZephyrFuel, AeroPump
           \end{itemize}
           \item SP: Willingness to Pay Extra for Green Fuel
           \begin{itemize}
               \item Attributes: Yes, No (cheapest preferred)
           \end{itemize}
           \item SP: Price Sensitivity for Fuel
           \begin{itemize}
               \item Attributes: Always cheapest, Rather cheapest, Price is irrelevant
           \end{itemize}
       \end{enumerate}
       \item Charging Station (in public)
       \begin{enumerate}
           \item MP: Preferred Charging Network
           \begin{itemize}
               \item Attributes: ChargeSwift, EcoPulse Energy, VoltRise Charging, AmpFlow Solutions, ZapGrid Power
           \end{itemize}
           \item SP: Preferred type of Charging while traveling
           \begin{itemize}
               \item Attributes: AC, DC, HPC
           \end{itemize}
           \item SP: Preferred type of Charging when being at everyday points (e.g., work, grocery, restaurant)
           \begin{itemize}
               \item Attributes: AC, DC, HPC
           \end{itemize}
           \item MP: Charging Station Amenities
           \begin{itemize}
               \item Attributes: On-site amenities (Restaurant/cafes), Wi-Fi availability, Seating area, Restroom facilities
           \end{itemize}
       \end{enumerate}
       \item Grocery Shopping
       \begin{enumerate}
           \item MP: Preferred Supermarket Chains
           \begin{itemize}
               \item Attributes: MarketMingle, FreshFare Hub, GreenGroove Stores, BasketBounty Markets, PantryPulse Retail
           \end{itemize}
           \item SP: Preference for Local Markets/Farms or Supermarket
           \begin{itemize}
               \item Attributes: Local Markets/Farms, Supermarket
           \end{itemize}
       \end{enumerate}
   \end{enumerate}
   \item Navigation and Routing
   \begin{enumerate}
       \item Routing
       \begin{enumerate}
           \item MP: Avoidance of Specific Road Types
           \begin{itemize}
               \item Attributes: Highways, Toll roads, Unpaved roads
           \end{itemize}
           \item SP: Priority for Shortest Time or Shortest Distance
           \begin{itemize}
               \item Attributes: Shortest Time, Shortest Distance
           \end{itemize}
           \item SP: Tolerance for Traffic
           \begin{itemize}
               \item Attributes: Low, Medium, High
           \end{itemize}
       \end{enumerate}
       \item Traffic and Conditions
       \begin{enumerate}
           \item SP: Traffic Information Source Preferences
           \begin{itemize}
               \item Attributes: In-car system, NavFlow Updates, RouteWatch Alerts, TrafficTrendz Insights
           \end{itemize}
           \item SP: Willingness to Take Longer Route to Avoid Traffic
           \begin{itemize}
               \item Attributes: Yes, No (traffic tolerated for fastest route)
           \end{itemize}
       \end{enumerate}
       \item Parking
       \begin{enumerate}
           \item SP: Preferred Parking Type
           \begin{itemize}
               \item Attributes: On-street, Off-street, Parking-house
           \end{itemize}
           \item SP: Price Sensitivity for Paid Parking
           \begin{itemize}
               \item Attributes: Always considers price first, Sometimes considers price, Never considers price
           \end{itemize}
           \item SP: Distance Willing to Walk from Parking to Destination
           \begin{itemize}
               \item Attributes: less than 5 min (accepting possible higher cost), less than 10 min (accepting possible higher cost), not relevant (closest with low cost)
           \end{itemize}
           \item SP: Preference for Covered Parking
           \begin{itemize}
               \item Attributes: Yes, Indifferent to Covered Parking
           \end{itemize}
           \item SP: Need for Handicapped Accessible Parking
           \begin{itemize}
               \item Attributes: Yes
           \end{itemize}
           \item SP: Preference for Parking with Security
           \begin{itemize}
               \item Attributes: Yes, Indifferent to Parking Security
           \end{itemize}
       \end{enumerate}
   \end{enumerate}
   \item Vehicle Settings and Comfort
   \begin{enumerate}
       \item Climate Control
       \begin{enumerate}
           \item SP: Preferred Temperature
           \begin{itemize}
               \item Attributes: 18 degree Celsius, 19 degree Celsius, 20 degree Celsius, 21 degree Celsius, 22 degree Celsius, 23 degree Celsius, 24 degree Celsius, 25 degree Celsius
           \end{itemize}
           \item SP: Fan Speed Preferences
           \begin{itemize}
               \item Attributes: Low, Medium, High
           \end{itemize}
           \item SP: Airflow Direction Preferences
           \begin{itemize}
               \item Attributes: Face, Feet, Centric, Combined
           \end{itemize}
           \item SP: Seat Heating Preferences
           \begin{itemize}
               \item Attributes: Low, Medium, High
           \end{itemize}
       \end{enumerate}
       \item Lighting and Ambience
       \begin{enumerate}
           \item SP: Interior Lighting Brightness Preferences
           \begin{itemize}
               \item Attributes: Low, Medium, High
           \end{itemize}
           \item SP: Interior Lighting Ambient Preferences
           \begin{itemize}
               \item Attributes: Warm, Cool
           \end{itemize}
           \item MP: Interior Lightning Color Preferences
           \begin{itemize}
               \item Attributes: Red, Blue, Green, Yellow, White, Pink
           \end{itemize}
       \end{enumerate}
   \end{enumerate}
   
   \item Entertainment and Media
   \begin{enumerate}
       \item Music
       \begin{enumerate}
           \item MP: Favorite Genres
           \begin{itemize}
               \item Attributes: Pop, Rock, Jazz, Classical, Country, Rap
           \end{itemize}
           \item MP: Favorite Artists/Bands
           \begin{itemize}
               \item Attributes: Max Jettison (Pop), Melody Raven (Pop), Melvin Dunes (Jazz), Ludwig van Beatgroove (Classical), Wolfgang Amadeus Harmonix (Classical), Taylor Winds (Country/Pop), Ed Sherwood (Pop/Folk), TwoPacks (Rap)
           \end{itemize}
           \item MP: Favorite Songs
           \begin{itemize}
               \item Attributes: Envision by Jon Lemon (Rock), Dreamer's Canvas by Lenny Visionary (Folk), Jenny's Dance by Max Rythmo (Disco), Clasp My Soul by The Harmonic Five (Soul), Echoes of the Heart by Adeena (R\&B), Asphalt Anthems by Gritty Lyricist (Rap), Cosmic Verses by Nebula Rhymes (Hip-Hop/Rap)
           \end{itemize}
           \item SP: Preferred Music Streaming Service
           \begin{itemize}
               \item Attributes: SonicStream, MelodyMingle, TuneTorrent, HarmonyHive, RhythmRipple
           \end{itemize}
       \end{enumerate}
       \item Radio and Podcasts
       \begin{enumerate}
           \item SP: Preferred Radio Station
           \begin{itemize}
               \item Attributes: EchoWave FM, RhythmRise Radio, SonicSphere 101.5, VibeVault 88.3, HarmonyHaven 94.7
           \end{itemize}
           \item MP: Favorite Podcast Genres
           \begin{itemize}
               \item Attributes: News, Technology, Entertainment, Health, Science
           \end{itemize}
           \item MP: Favorite Podcast Shows
           \begin{itemize}
               \item Attributes: GlobalGlimpse News, ComedyCraze, ScienceSync, FantasyFrontier, WellnessWave
           \end{itemize}
           \item SP: General News Source
           \begin{itemize}
               \item Attributes: NewsNexus, WorldPulse, CurrentConnect, ReportRealm, InfoInsight
           \end{itemize}
       \end{enumerate}
   \end{enumerate}
\end{enumerate}

\section{Methodology: Preference Extraction}\label{sec:app_preference_extraction}

We define the LLM function for extracting user preferences as follows:

\vspace{-.7\baselineskip}
\begin{minted}[breaklines, fontsize=\small, bgcolor=gray!10]{json}
"type": "function",
"function": {
    "name": "extract_user_preference",
    "description": "A function that extracts personal preferences of the user ...",
    "parameters": "<nested parameter schema representing the hierarchical categories>"}
\end{minted}
\vspace{-1.5\baselineskip}

The parameter schema, defined using Pydantic, includes categories and their hierarchy. Below is a representative subset for the main category \texttt{Points of Interest}, sub-category \texttt{Restaurant}, and detail-category \texttt{Favourite Cuisine}:

\begin{minted}[breaklines, fontsize=\small, bgcolor=gray!10]{python}
class PreferencesFunctionOutput(BaseModel):
    points_of_interest: Optional[PointsOfInterest] = Field(default=None,
        description="The user's preferences in the category 'Points of Interest'.",)
    navigation_and_routing: Optional[NavAndRouting] = Field(...)
    ...

class PointsOfInterest(BaseModel):
    no_or_other_preference: ...
    restaurant: Optional[Restaurant] = Field(defualt=None, description="...")
    ...
    
class Restaurant(BaseModel):
    no_or_other_preference: ...
    favourite_cuisine: Optional[List[OutputFormat]] = Field(default=[], description="...", examples=["Italian", "Chinese", ...])
    ...
    
class OutputFormat(BaseModel):
    user_sentence_preference_revealed: Optional[str] = Field(default=None, description="user sentence where the user revealed the preference.")
    user_preference: Optional[str] = Field(default=None, description="The preference of the user.")
\end{minted}

Each category is represented as a parameter with a type, default value, description, and optional examples. The nested schema represents the relationship of the categories. As every parameter is \texttt{Optional}, the LLM is not forced to extract a preference for every parameter within that category. We found that including the parameter \verb|no_or_other_preference| within the sub- and detail categories reduces over-extraction, as the LLM must actively decide not to place a preference there if it intends to extract one. Through the \texttt{Output Format}, we can see, that the LLM should not only extract the preference itself, but also the sentence where the user revealed the preference.

\section{Additional Experiment Results}
\label{sec:app_add_exp_results}

\subsection{Confusion Matrices of Preference Extraction}\label{subsec:app_cm_detail_cat_in}

\begin{figure*}[t]
    \centering
    \includegraphics[width=\linewidth]{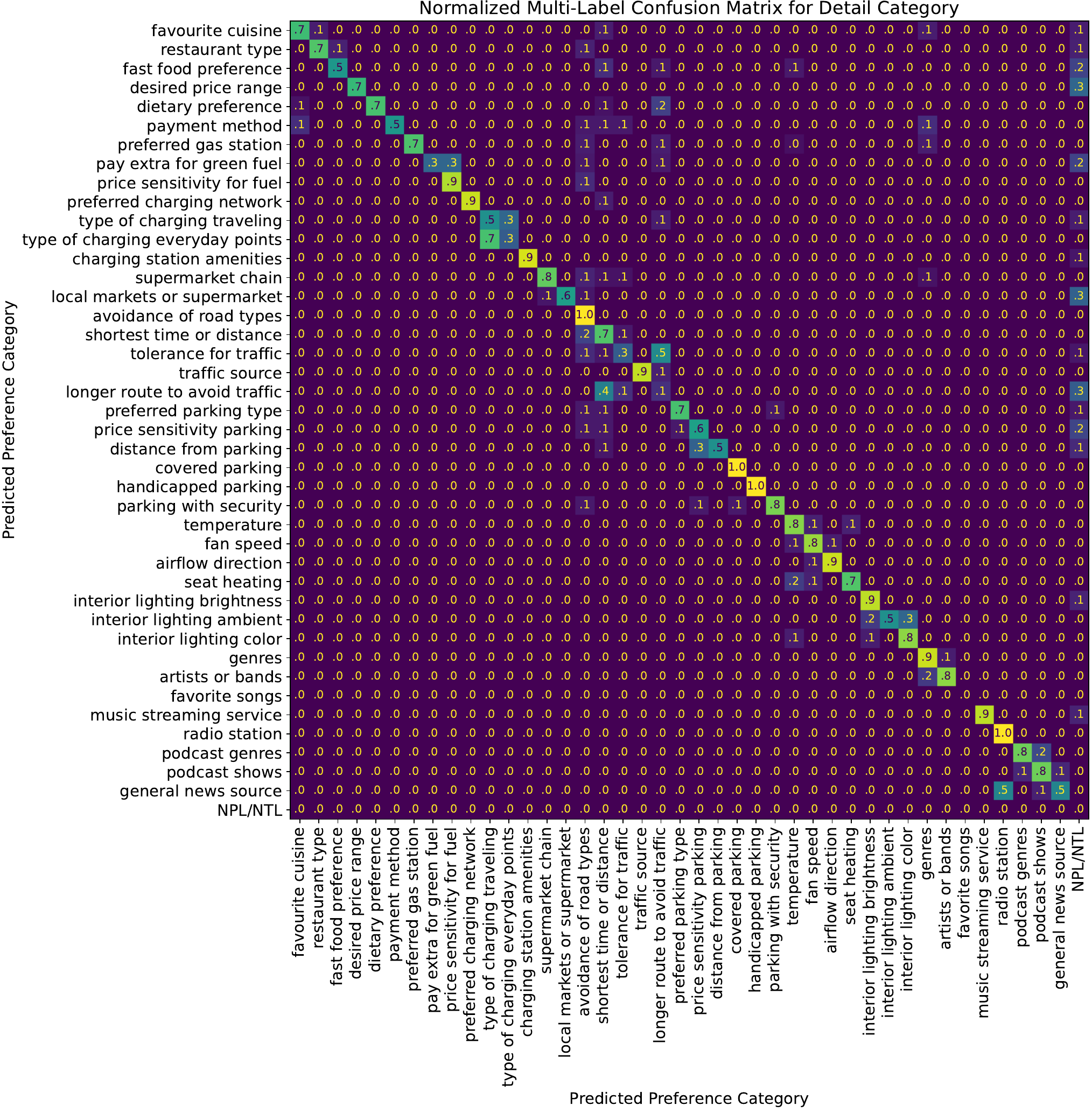}
    \caption{Multi-Label confusion matrix \cite{MultiLabelConfusionMatrix}, normalized across the rows, on the detail category level for the In-Schema experiments (refer to Section \ref{subsec:exp_pref_extraction}). The last row represents data points with no true label (NTL), while the last column represents data points with no predicted label (NPL).}
    \label{fig:cm_detail}
\end{figure*}

Figure \ref{fig:cm_detail} shows the multi-label confusion matrix on the detail category level for the In-Schema experiment (refer to Section \ref{subsec:exp_pref_extraction}).

The strong diagonal in the confusion matrix indicates that the extraction process reliably adheres to the category schema. Most incorrect extractions occur in semantically related categories. After manual analysis, we found that the increased misclassifications in the detail category 'avoidance of specific roadtypes', 'shortest time or distance', and 'tolerance for traffic' are mostly due to the dataset. During dataset generation, an extra preference is occasionally included in the user utterances within these categories, as in-car conversations often evolve toward these topics naturally.

\begin{figure}[ht]
    \centering
    \includegraphics[width=\linewidth]{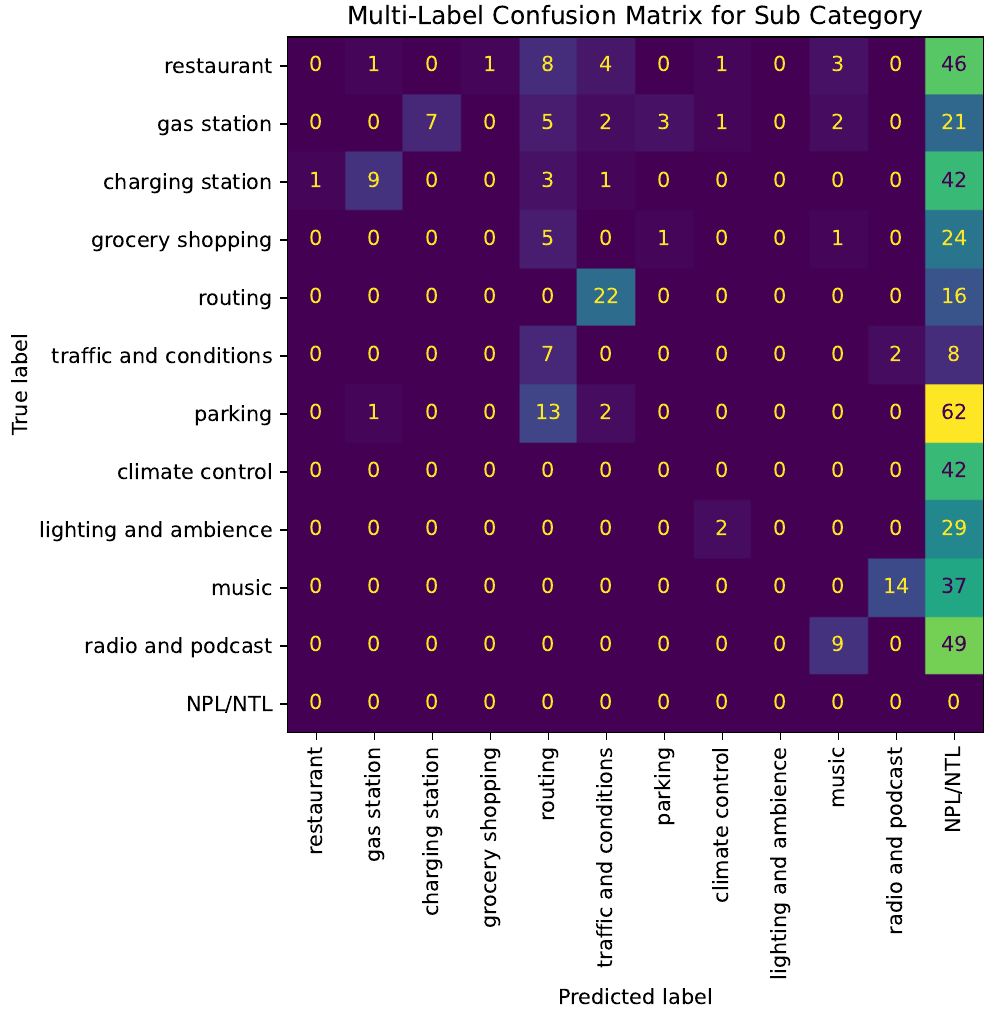}
    \caption{Multi-label confusion matrix \cite{MultiLabelConfusionMatrix} on the subcategory level for the Out-of-Schema experiment (refer to Section \ref{subsec:exp_pref_extraction}). The last row represents data points with no true label (NTL), while the last column represents data points with no predicted label (NPL). In this experiment, it is expected to have no predicted label for every data point.}
    \label{fig:cm_sub_cat_out}
\end{figure}

Figure \ref{fig:cm_sub_cat_out} shows the multi-label confusion matrix on the subcategory level for the Out-of-Schema experiment (refer to Section \ref{subsec:exp_pref_extraction}). As the category of the ground-truth preference is excluded in the schema for this experiment, we expect the system to perform no extraction. We see that we have few incorrect extractions when excluding semantically distinct categories such as 'Climate Control' ($0$ incorrect extraction), but significantly more if there is still a closely related category like in 'Music' and 'Radio and Podcast'. This indicates that the definition of clear and semantically distinct categories is key to a reliable category-bound extraction.

\end{document}